# Multi-Level Discovery of Deep Options

**Roy Fox**[*]
EECS Department
UC Berkeley
`royf@berkeley.edu`

**Sanjay Krishnan**[*]
EECS Department
UC Berkeley
`sanjaykrishnan@berkeley.edu`

**Ion Stoica**
EECS Department
UC Berkeley
`istoica@berkeley.edu`

**Ken Goldberg**
EECS and IEOR Departments
UC Berkeley
`goldberg@berkeley.edu`

## Abstract

Augmenting an agent's control with useful higher-level behaviors called *options* can greatly reduce the sample complexity of reinforcement learning, but manually designing options is infeasible in high-dimensional and abstract state spaces. While recent work has proposed several techniques for automated option discovery, they do not scale to multi-level hierarchies and to expressive representations such as deep networks. We present *Discovery of Deep Options (DDO)*, a policy-gradient algorithm that discovers parametrized options from a set of demonstration trajectories, and can be used recursively to discover additional levels of the hierarchy. The scalability of our approach to multi-level hierarchies stems from the decoupling of low-level option discovery from high-level meta-control policy learning, facilitated by under-parametrization of the high level. We demonstrate that using the discovered options to augment the action space of Deep Q-Network agents can accelerate learning by guiding exploration in tasks where random actions are unlikely to reach valuable states. We show that DDO is effective in adding options that accelerate learning in 4 out of 5 Atari RAM environments chosen in our experiments. We also show that DDO can discover structure in robot-assisted surgical videos and kinematics that match expert annotation with 72% accuracy.

## 1 Introduction

Control policies that perform long and intricate tasks often require high-dimensional parametrization. In principle, existing reinforcement learning (RL) algorithms can learn such policies, but often with prohibitive sample complexity. One approach is to augment the agent's controls with useful higher-level behaviors called *options* [1], each consisting of a control policy for one region of the state space, and a termination condition recognizing leaving that region. This augmentation naturally defines a hierarchical structure of high-level *meta-control* policies that invoke lower-level options to solve sub-tasks. This leads to a "divide and conquer" relationship between the levels, where each option can specialize in short-term planning over local state features, and the meta-control policy can specialize in long-term planning over slowly changing state features.

To apply RL in long-horizon tasks, such as driving and surgery, we need to consider *deep hierarchies* of *deep options* — policies that employ multi-level hierarchies of expressive options parametrized by deep neural networks. Since designing such structures manually is infeasible, we need algorithms to discover them. Despite recent results in option discovery, some proposed techniques do not generalize well to multi-level hierarchies [2, 3, 4], while others are inefficient for learning expressive representations [5, 6, 7, 8].

---
[*]Equal contribution

We introduce *Discovery of Deep Options (DDO)*, an algorithm for efficiently discovering deep hierarchies of deep options. DDO is a policy-gradient algorithm that discovers parametrized options from a set of demonstration trajectories (sequences of states and actions) provided either by a supervisor or by roll-outs of previously learned policies. These demonstrations need not be given by an optimal agent, but it is assumed that they are informative of the preferred actions to take in each visited state, and are not just random walks.

The basis for this algorithm is Hierarchical Behavioral Cloning (HBC), where the agent tries to learn a hierarchical policy that matches the demonstrated behavior. Given a set of trajectories, the algorithm discovers a fixed, predetermined number of options that are most likely to generate the observed trajectories. Since an option is represented by both a control policy and a termination condition, our algorithm simultaneously (1) infers option boundaries in demonstrations which segment trajectories into different control regimes, (2) infers the meta-control policy for selecting options as a mapping of segments to the option that likely generated them, and (3) learns a control policy for each option, which can be interpreted as a soft clustering where the centroids correspond to prototypical behaviors of the agent.

The algorithm repeats for each additional level of the hierarchy. The scalability of our approach to multi-level hierarchies stems from the decoupling of low-level option discovery from high-level meta-control policy learning. The true meta-control policy is often too richly parametrized, and must be simplified to facilitate low-level option discovery. Each higher level of the hierarchy can then be discovered separately, building on established lower-level options, without involving any unlearned higher levels. We show that DDO is effective in adding to the hierarchy more levels that can further accelerate learning.

We demonstrate that using the discovered options in Deep Q-Network agents can accelerate reinforcement learning by guiding exploration. Useful options are persistent control loops that lead to valuable states, from which exploration can continue. When these states are unlikely to be reached by random actions, using options allows an $\epsilon$-greedy exploring agent to nevertheless reach these valuable states by suspending exploration until the option terminates.

## 2 Related Work

The field of hierarchical reinforcement learning has a long history [9, 1, 10], and has also been applied in robotics [11, 12, 13] and in the analysis of biological systems [14, 15, 16, 17, 18]. Early work in hierarchical control demonstrated the advantages of hierarchical structures by handcrafting hierarchical policies [19] and by learning them given various manual specifications: state abstractions [20, 21, 22, 23], a set of waypoints [24], low-level skills [25, 26, 27], a set of finite-state meta-controllers [28], a set of subgoals [1, 29], or intrinsic reward [2]. Since then, the focus of research has shifted towards discovery of the hierarchical structure itself, by: trading off value with description length [30], identifying transitional states [31, 32, 33, 34, 6], inference from demonstrations [8, 35, 5, 12], iteratively expanding the set of solvable initial states [36, 11], policy gradient [37], trading off value with informational constraints [38, 39, 40, 41], active learning [7], or recently value-function approximation [4, 3, 42].

Most previous work in hierarchy discovery focused on a 2-level hierarchy, and its extension to multi-level hierarchies is not always straightforward. Naively, the number of different contexts that the $d$ highest levels can provide to lower levels grows exponentially with $d$. Discovery of multi-level hierarchies was shown to be possible via particle filters [8], however we are seeking more expressive representations. We show that this "curse of dimensionality" of multi-level meta-control can be avoided if, during the option discovery phase, we make a modeling assumption that under-parametrizes the higher-level model — e.g. fixes the meta-control policy. This decouples the lower-level discovery from the levels above it. Empirically, we find that this assumption does not sacrifice the quality of the discovered options (Section 5.1).

Our work is most related to [43], who use a similar generative model, originally introduced by [8] as an Abstract Hidden Markov Model, and learn its parameters via the Expectation-Maximization (EM) algorithm. EM applies the same forward-backward E-step as our Expectation-Gradient al-gorithm (Section 4.2) to compute marginal posteriors, but uses them for a complete optimization M-step over the options and the meta-control policies. This optimization is infeasible for expressive representations, as well as for multi-level hierarchies. Gradient-descent algorithms for value-function approximation with deep networks has been used to train hierarchical policies [2, 3], using a Universal Value Function Approximator [44]. These methods learn the entire hierarchy simultaneously, and may be inefficient for multi-level hierarchies.

## 3 Preliminaries

### 3.1 Markov Decision Processes

We consider a discrete-time discounted Markov Decision Process (MDP), described by a 6-tuple $\langle \mathcal{S}, \mathcal{A}, p_0, p, r, \gamma \rangle$, where $\mathcal{S}$ denotes the state space, $\mathcal{A}$ the action space, $p_0(s_0)$ the initial state distribution, $p(s_{t+1}|s_t, a_t)$ the state transition distribution, $r(s_t, a_t) \in \mathbb{R}$ the reward function, and $\gamma \in [0, 1)$ the discount factor. A policy $\pi(a_t|s_t)$ defines a conditional probability distribution over actions given the state. A trajectory is defined as a sequence of states and actions $\xi = (s_0, a_0, s_1, \ldots, s_T)$ of a given length $T$. In a given MDP, a policy $\pi$ induces the distribution over trajectories:

$$\mathbb{P}_\pi(\xi) = p_0(s_0) \prod_{t=0}^{T-1} \pi(a_t|s_t) p(s_{t+1}|s_t, a_t).$$

The *return* of a policy is its expected total discounted reward over trajectories

$$V_\pi = \mathbb{E}_{\xi \sim \mathbb{P}_\pi} \left[ \sum_{t=0}^{T-1} \gamma^t r(s_t, a_t) \right].$$

### 3.2 Reinforcement Learning

In many real-world domains, we seek a policy $\pi$ that achieves high return despite the state transition distribution $p$ being initially unknown. One successful approach to this problem is Q-learning [45, 46], a reinforcement learning algorithm for estimating the maximum return $Q(s, a)$, which is the expected total discounted reward obtained by taking the action $a$ in the state $s$, and thereafter following the optimal policy.

Q-learning is off-policy, in that it samples a state $s_i$, action $a_i$, reward $r_i = r(s_i, a_i)$ and next state $s'_i \sim p(\cdot|s_i, a_i)$ using some suboptimal exploration policy. Based on these samples, Q-learning then updates its estimate of $Q(s_i, a_i)$ greedily towards the one-step backward estimate

$$y_i = r_i + \gamma \max_{a'} Q(s'_i, a').$$

When $Q_\theta$ is represented by a Deep Q-Network (DQN) [47], the update is a gradient step to reduce the total square *Bellman error* over a batch of $N$ samples:

$$L[\theta] = -\sum_{i=1}^{N} (y_i - Q_\theta(s_i, a_i))^2$$

$$\theta \leftarrow \theta + \alpha \nabla_\theta L[\theta] = \theta + \alpha \sum_{i=1}^{N} (y_i - Q_\theta(s_i, a_i)) \nabla_\theta Q_\theta(s_i, a_i),$$

where $\alpha$ is the learning rate, and $y_i$ is taken as independent of $\theta$.

### 3.3 Imitation Learning

Imitation learning (IL) is an alternative to the reinforcement learning setting, in which the policy is learned from demonstrations of expert behavior rather than from the reward signal of an environment. In the behavioral cloning (BC) setting, the goal is to match the expert policy as closely as possible.

We can formalize this setting as estimation of the parameters of a generative model. Suppose that a demonstration trajectory $\xi = (s_0, a_0, s_1, \ldots, s_T)$ is generated by an unknown policy $\pi$ according to the following generative model:

Initialize $t \leftarrow 0$, $s_0 \sim p_0$
**for** $t \leftarrow 0, \ldots, T-1$ **do**
    Draw $a_t \sim \pi(\cdot|s_t)$
    Draw $s_{t+1} \sim p(\cdot|s_t, a_t)$

We can define a parametrized set of policies $\pi_\theta$ and find the parameters that maximize the log-likelihood

$$L[\theta; \xi] = \log p_0(s_0) + \sum_{t=0}^{T-1} \log(\pi_\theta(a_t|s_t) p(s_{t+1}|s_t, a_t)).$$

It is interesting to note that the dynamics factor out of this optimization problem, simplifying it to

$$\arg\max_{\theta \in \Theta} L[\theta; \xi] = \arg\max_{\theta \in \Theta} \sum_{t=0}^{T-1} \log \pi_\theta(a_t|s_t).$$

For differentiable parametrizations the gradient update is

$$\theta \leftarrow \theta + \alpha \sum_{t=0}^{T-1} \nabla_\theta \log \pi_\theta(a_t|s_t).$$

### 3.4 The Options Framework

Control policies that perform long and intricate tasks often require very high-dimensional parameterization, which can take many trajectories to learn. The options framework is a hierarchical policy structure that can mitigate this sample complexity by breaking down the policy into simpler skills called *options* [1]. An option represent a lower-level control primitive that can be invoked by the *meta-control* policy at a higher-level of the hierarchy, in order to perform a certain subtask. Formally, an option $h$ is described by a triplet $\langle \mathcal{I}_h, \pi_h, \psi_h \rangle$, where $\mathcal{I}_h \subseteq \mathcal{S}$ denotes the initiation set, $\pi_h(a_t|s_t)$ the control policy, and $\psi_h(s_t) \in [0, 1]$ the termination policy. When the process reaches a state $s \in \mathcal{I}_h$, the option $h$ can be invoked to run the policy $\pi_h$. After each action is taken and the next state $s'$ is reached, the option $h$ terminates with probability $\psi_h(s')$ and returns control up the hierarchy to its invoking level.

The options framework enables multi-level hierarchies to be formed by allowing options to invoke other options. A higher-level meta-control policy is defined by augmenting its action space $\mathcal{A}$ with the set $\mathcal{H}$ of all lower-level options. In this paper we do not explicitly consider initiation sets. Instead, any higher-level policy $\eta(h|s)$ should assign low probability to a lower-level option $h$ outside the region of state space in which $h$ is effective.

## 4 Discovery of Deep Options

In this section, we present *Discovery of Deep Options (DDO)*, a policy-gradient algorithm that discovers parametrized options from a set of demonstration trajectories (sequences of states and actions). The option parameters are inferred by fitting a generative model to the observed trajectories. These demonstrations need not be given by an optimal agent — the demonstrator can be a human who is not an expert, or a partially trained algorithmic controller. Our approach only assumes that the trajectories are informative of the preference of actions to take in each visited state, and that these preferences can be represented in a hierarchical structure (see Section C).

### 4.1 Imitation Learning for Option Discovery

We generalize standard IL to hierarchical control, by introducing Hierarchical Behavioral Cloning (HBC). In HBC, the meta-control signals that form the hierarchy are unobservable, latent variables of the generative model, that must be inferred.

Consider a trajectory $\xi = (s_0, a_0, s_1, \ldots, s_T)$ that is generated by a two-level hierarchy. The low level implements a set $\mathcal{H}$ of options $\langle \pi_h, \psi_h \rangle_{h \in \mathcal{H}}$. The high level implements a meta-control policy $\eta(h_t|s_t)$ that repeatedly chooses an option $h_t \sim \eta(\cdot|s_t)$ given the current state, and runs it until termination. Our hierarchical generative model is:

```
Initialize t ← 0, s_0 ~ p_0, b_0 ← 1
for t ← 0, ..., T − 1 do
    if b_t = 1 then
        Draw h_t ~ η(·|s_t)
    else   // b_t = 0
        Set h_t ← h_{t−1}
    Draw a_t ~ π_{h_t}(·|s_t)
    Draw s_{t+1} ~ p(·|s_t, a_t)
    Draw b_{t+1} ~ Ber(ψ_{h_t}(s_{t+1}))
```

### 4.2 Expectation-Gradient Algorithm

We denote by $\theta$ the vector of parameters for $\pi_h$, $\psi_h$ and $\eta$. For example, $\theta$ can be the weights and biases of a feed-forward network that computes these probabilities. This generic notation allows us the flexibility of a completely separate network for the meta-control policy and for each option, $\theta = (\theta_\eta, (\theta_h)_{h \in \mathcal{H}})$, or the efficiency of sharing some of the parameters between options, similarly to a Universal Value Function Approximator [44].

We want to find the $\theta \in \Theta$ that maximizes the log-likelihood assigned to a given dataset of trajectories. The likelihood of a trajectory depends on the latent sequence $\zeta = (b_0, h_0, b_1, h_1, \ldots, h_{T-1})$ of meta-actions and termination indicators, and in order to use a gradient-based optimization method we rewrite the gradient using the following *EG-trick*:

$$\nabla_\theta L[\theta; \xi] = \nabla_\theta \log \mathbb{P}_\theta(\xi) = \frac{1}{\mathbb{P}_\theta(\xi)} \sum_{\zeta \in (\{0,1\} \times \mathcal{H})^T} \nabla_\theta \mathbb{P}_\theta(\zeta, \xi)$$

$$= \sum_\zeta \frac{\mathbb{P}_\theta(\zeta, \xi)}{\mathbb{P}_\theta(\xi)} \nabla_\theta \log \mathbb{P}_\theta(\zeta, \xi) = \mathbb{E}_{\zeta|\xi;\theta}[\nabla_\theta \log \mathbb{P}_\theta(\zeta, \xi)],$$

which is the so-called *Expectation-Gradient* method [48, 49].

The generative model in the previous section implies the likelihood

$$\mathbb{P}_\theta(\zeta, \xi) = p_0(s_0) \delta_{b_0=1} \eta(h_0|s_0) \prod_{t=1}^{T-1} \mathbb{P}_\theta(b_t, h_t|h_{t-1}, s_t) \prod_{t=0}^{T-1} \pi_{h_t}(a_t|s_t) p(s_{t+1}|s_t, a_t),$$

with

$$\mathbb{P}_\theta(b_t=1, h_t|h_{t-1}, s_t) = \psi_{h_{t-1}}(s_t) \eta(h_t|s_t)$$
$$\mathbb{P}_\theta(b_t=0, h_t|h_{t-1}, s_t) = (1 - \psi_{h_{t-1}}(s_t)) \delta_{h_t=h_{t-1}}.$$

Applying the EG-trick and ignoring the terms that do not depend on $\theta$, we can simplify the gradient to:

$$\nabla_\theta L[\theta; \xi] = \mathbb{E}_{\zeta|\xi;\theta}\left[\nabla_\theta \log \eta(h_0|s_0) + \sum_{t=1}^{T-1} \nabla_\theta \log \mathbb{P}_\theta(b_t, h_t|h_{t-1}, s_t) + \sum_{t=0}^{T-1} \nabla_\theta \log \pi_{h_t}(a_t|s_t)\right].$$

The log-likelihood gradient can therefore be computed as the sum of the log-probability gradients of the various parameterized networks, weighed by the marginal posteriors

$$u_t(h) = \mathbb{P}_\theta(h_t=h|\xi)$$
$$v_t(h) = \mathbb{P}_\theta(b_t=1, h_t=h|\xi)$$
$$w_t(h) = \mathbb{P}_\theta(h_t=h, b_{t+1}=0|\xi).$$

In the Expectation-Gradient algorithm, the E-step computes $u$, $v$ and $w$, and the G-step updates the parameter with a gradient step, namely

$$\nabla_\theta L[\theta; \xi] = \sum_{h \in \mathcal{H}} \left( \sum_{t=0}^{T-1} \left( v_t(h) \nabla_\theta \log \eta(h|s_t) + u_t(h) \nabla_\theta \log \pi_h(a_t|s_t) \right) \right.$$

$$\left. + \sum_{t=0}^{T-2} \left( (u_t(h) - w_t(h)) \nabla_\theta \log \psi_h(s_{t+1}) + w_t(h) \nabla_\theta \log(1 - \psi_h(s_{t+1})) \right) \right).$$

These equations lead to the natural iterative algorithm. In each iteration, the marginal posteriors $u$, $v$ and $w$ can be computed with a forward-backward message-passing algorithm similar to Baum-Welch [50], with time complexity $O(|\mathcal{H}|^2 T)$. Importantly, this algorithm can be performed without any knowledge of the state dynamics. The details of this computation are given in the supplementary material. Then, the computed posteriors can be used in a gradient descent algorithm to update the parameters:

$$\theta \leftarrow \theta + \alpha \sum_i \nabla_\theta L[\theta; \xi_i].$$

This update can be made stochastic using a single trajectory, uniformly chosen from the demonstration dataset, to perform each update.

Intuitively, the algorithm attempts to jointly optimize three objectives:

- Infer the option boundaries in which $b = 1$ appears likely relative to $b = 0$, as given by $(u - w)$ and $w$ respectively — this segments the trajectory into regimes where we expect $h$ to persist and employ the same control law; in the G-step we reduce the cross-entropy loss between the unnormalized distribution $(w, u - w)$ and the termination indicator $\psi_h$;

- Infer the option selection after a switch, given by $v$; in the G-step we reduce the cross-entropy loss between that distribution, weighted by the probability of a switch, and the meta-control policy $\eta$; and

- Reduce the cross-entropy loss between the empirical action distribution, weighted by the probability for $h$, and the control policy $\pi_h$.

This can be interpreted as a form of soft clustering. The data points are one-hot representations of each $a_t$ in the space of distributions over actions. Each time-step $t$ is assigned to option $h$ with probability $u_t(h)$, forming a soft clustering of data points. The G-step directly minimizes the KL-divergence of the control policy $\pi_h$ from the weighted centroid of the corresponding cluster.

### 4.3 Deeper Hierarchies

Our ultimate goal is to use the algorithm presented here to discover a multi-level hierarchical structure — the key insight being that the problem is recursive in nature. A $D$-level hierarchy can be viewed as a 2-level hierarchy, in which the "high level" has a $(D - 1)$-level hierarchical structure. The challenge is the coupling between the levels; namely, the value of a set of options is determined by its usefulness for meta-control [39], while the value of a meta-control policy depends on which options are available. This potentially leads to an exponential growth in the size of the latent variables required for inference. The available data may be insufficient to learn a policy so expressive.

We can avoid this problem by using a simplified parametrization for the intermediate meta-control policy $\eta_d$ used when discovering level-$d$ options. In the extreme, we can fix a uniform meta-control policy that chooses each option with probability $1/|\mathcal{H}_d|$. Discovery of the entire hierarchy can now proceed recursively from the lowest level upward: level-$d$ options can invoke already-discovered lower-level options; and are discovered in the context of a simplified level-$d$ meta-control policy, decoupled from higher-level complexity. One of the contributions of this work is to demonstrate that, perhaps counter-intuitively, this assumption does not sacrifice too much during option discovery. An informative meta-control policy would serve as a prior on the assignment of demonstration segments to the options that generated them, but with sufficient data this assignment can also be inferred from the low-level model, purely based on the likelihood of each segment to be generated by each option.

We use the following algorithm to iteratively discover a hierarchy of $D$ levels, each level $d$ consisting of $k_d$ options:

**for** $d = 1, \ldots, D - 1$ **do**
    Initialize a set of options $\mathcal{H}_d = \{h_{d,1}, \ldots, h_{d,k_d}\}$
    DDO: train options $\langle \pi_h, \psi_h \rangle_{h \in \mathcal{H}_d}$ with $\eta_d$ fixed
    Augment action space $\mathcal{A} \leftarrow \mathcal{A} \cup \mathcal{H}_d$
Use RL algorithm to train high-level policy

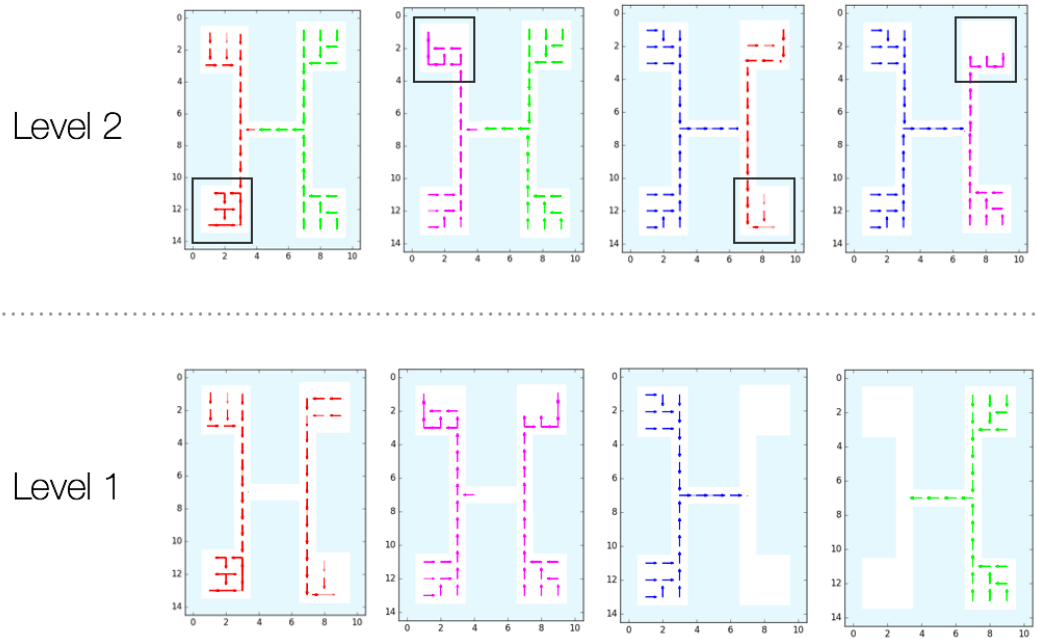

Figure 1: Trajectories generated by options in a hierarchy discovered by DDO. Each level has 4 options, where level 2 builds on the level 1, which in turn uses atomic actions. In the lower level, two of the options move the agent between the upper and lower rooms, while the other two options move it from one side to the other. In the higher level, each option takes the agent to a specific room. The lower-level options are color-coded to show how they are composed into the higher-level options.

## 5 Experiments

We present an empirical study of DDO. Our results suggest that DDO can discover options that accelerates reinforcement learning. We explore two different scenarios: (**Supervised**) given a supervisor who demonstrates a few times how to perform a task, show that the discovered options are useful for accelerating reinforcement learning on the same task; (**Exploration**) apply reinforcement learning for $T$ episodes, sample trajectories from the current best policy, and augment the action space with the discovered options for the remaining $T'$ episodes. The first set of experiments illustrates DDO on a series of GridWorld domains. Then, we show how the Expectation-Gradient can scale to more challenging Atari RAM domains. Finally, we show that DDO can be used to identify visuomotor primitives in surgical data.

### 5.1 Four Rooms GridWorld

We study a simple four-room domain (Figure 1). On a $15 \times 11$ grid, the agent can move in four directions; moving into a wall has no effect. To simulate environment noise, we replace the agent's action with a random one with probability $0.3$. An observable apple is spawned in a random location in one of the rooms. Upon taking the apple, the agent gets a unit reward and the apple is re-spawned.

We use the following notation to describe the different ways we can parametrize option discovery: A a baseline of only using atomic actions; H1u discovering a single level of options where the higher-level is parametrized by a uniform distribution; H1s discovering a single level of options where the higher-level is parametrized by an multi-layer perceptron (MLP); H2u and H2s are the two-level counterparts of H1u and H1s, respectively. All of these discovered options are used in an RL phase to augment the action space of a high-level global policy.

**Supervised Setting.** We use Value Iteration to compute the optimal policy, and then use this policy as a supervisor to generate 50 trajectories of length 1000. All policies, whether for control, meta-control or termination, are parametrized by a MLP, with a single two-node hidden layer, and $\tanh$ activation functions. The MLP's input consists of the full state (agent and apple locations), and the output is computed by a $\mathrm{softmax}$ function over the MLP output vector, which has length $|\mathcal{A}|$ for control policies and two for termination.

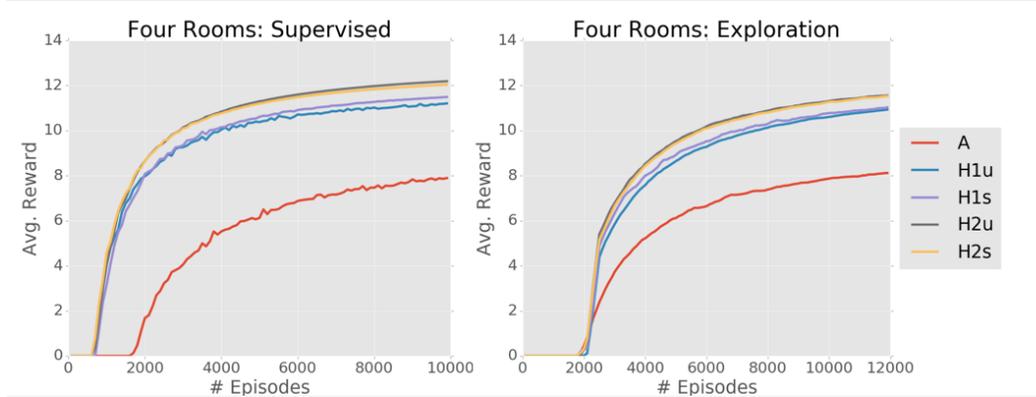

Figure 2: 15-trial mean reward for the Supervised and Exploration problem settings when running DQN with no options (A), low-level options (H1u) and lower- and higher-level options (H2u) augmenting the action space. The options discovered by DDO can accelerate learning, since they benefit from not being interrupted by random exploration.

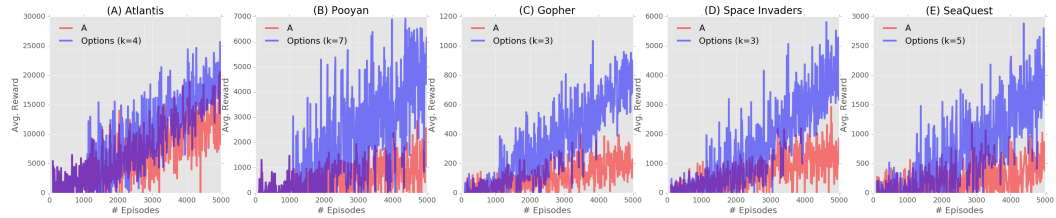

Figure 3: Atari RAM Games: Average reward computed from 50 rollouts when running DQN with atomic actions for 1000 episodes, then generating 100 trajectories from greedy policy, from which DDO discovers options in a 2-level hierarchy. DQN is restarted with action space augmented by these options, which accelerates learning in comparison to running DQN with atomic actions for 5000 episodes. Results suggest significant improvements in 4 out of 5 domains.

The options corresponding to H2u are visualized in Figure 1 by trajectories generated using each option from a random initial state until termination. At the first level, two of the discovered options move the agent between the upper and lower rooms, and two move it from one side to the other. At the second level, the discovered options aggregate these primitives into higher-level behaviors that move the agent from any initial location to a specific room.

**Impact of options and hierarchy depth.** To evaluate the quality of the discovered options, we train a Deep Q-Network (DQN) with the same MLP architecture, and action space augmented by the options. The exploration is $\epsilon$-greedy with $\epsilon = 0.2$ and the discount factor is $\gamma = 0.9$. Figure 2 shows the average reward in 15 of the algorithm's runs in the Supervised and Exploration experimental settings. The results illustrate that augmenting the action space with options can significantly accelerate learning. Note that the options learned with a two-level hierarchy (H2u) provide significant benefits over the options learned only with a single-level hierarchy H1u. The hierarchical approaches achieve roughly the same average reward after 1000 episodes as A does after 5000 episodes.

**Impact of policy parametrization.** To evaluate the effect of meta-control policy parametrization, we also compare the rewards during DQN reinforcement learning with options discovered with MLP meta-control policies (H1s and H2s). Our empirical results suggest that less expressive parametrization of the meta-control policy does not significantly hurt the performance, and in some cases can even provide a benefit (Figure 2). This is highly important, because the high sample complexity of jointly training all levels of a deep hierarchy necessitates simplifying the meta-control policy — which would otherwise be represented by a one level shallower hierarchy. We conjecture that the reason for the improved performance of the less expressive model is that more complex parametrization of the meta-control policy increases the prevalence of local optima in the inference problem, which may lead to worse options.

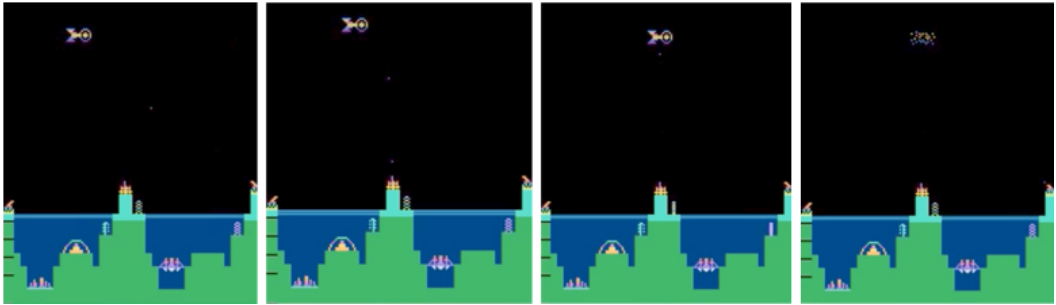

Figure 4: Frames sampled from a demonstration trajectory assigned to one of the primitives learned from DDO.

**Exploration Setting.** Finally, we demonstrate that options can also be useful when discovered from self-demonstrations by a partially trained agent, rather than by an expert supervisor. We run the same DQN as above, with only atomic actions, for 2000 episodes. We then use the greedy policy for the learned $Q$-function to generate 100 trajectories. We reset the $Q$-function (except for the baseline A), and run DQN again with the augmented action space. Figure 2 illustrates that even when these options are not discovered from demonstrations of optimal behavior, they are useful in accelerating reinforcement learning. The reason is that options are policy fragments that have been discovered to lead to interesting states, and therefore benefit from not being interrupted by random exploration.

## 5.2 Atari RAM Games

The RAM variant of the popular Atari Deep Reinforcement Learning domains considers a game-playing agent which is given not the screen, but rather the RAM state of the Atari machine. This RAM state is a 128-byte vector that completely determines the state of the game, and can be encoded in one-hot representation as $s \in \mathbb{R}^{128 \times 256}$. The RAM state-space illustrates the power of an automated option discovery framework, as it would be infeasible to manually code options without carefully understanding the game's memory structure. With a discovery algorithm, we have a general-purpose approach to learn in this environment.

All policies are parametrized with a deep network. There are three dense layers, each with $\tanh$ activations, and the output distribution is a $\mathrm{softmax}$ of the last layer, which has length $|\mathcal{A}|$ for control policies and two for termination. We use a single level of options, with the number of options tuned to optimize performance, and given in Figure 3.

For each game, we first run the DQN for 1000 episodes, and then generate 100 trajectories from the greedy policy, and use them to discover options with DDO. The DQN has the same architecture, using $\epsilon$-greedy exploration for 1000 episodes with $\epsilon = 0.05$ and discount factor $\gamma = 0.85$ (similar to the parameters used in [51]). Finally, we augment the action space with the discovered options and rerun DQN for 4000 episodes. We compare this to the baseline of running DQN for 5000 episodes with actions only.

Figure 3 plots the estimate value, averaged over 50 trials, of the learned policies for five Atari games: Atlantis, Pooyan, Gopher, Space Invaders, and Sea Quest. In four out of five games, we see a significant acceleration learning. The relative improvements are the largest for the three hardest domains: Gopher, Sea Quest, and Space Invaders. It is promising that DDO offers such an advantage where other methods struggle. Figure 4 shows four frames from one of the options discovered by DDO for the Atlantis game. The option appears to identify an incoming alien and determine when to fire the gun, terminating when the alien is destroyed. As in the GridWorld experiments, the options are policy fragments that have been discovered to lead to high-value states, and therefore benefit from not being interrupted by random exploration.

### 5.3 Segmentation of Robotic-Assisted Surgery

In this section, we illustrate the wide applicability of the DDO framework by applying it to human demonstrations in a robotic domain. We apply DDO to long robotic trajectories (e.g. 3 minutes) demonstrating an intricate task, and discover options for useful subtasks, as well as segmentation of the demonstrations into semantic units. The JIGSAWS dataset consists of surgical robot trajectories of human surgeons performing training procedures [52]. The dataset was captured using the da Vinci Surgical System from eight surgeons with different skill levels, performing five repetitions each of *needle passing*, *suturing*, and *knot tying*. This dataset consists of videos and kinematic data of the robot arms, and is annotated by experts identifying the activity occurring in each frame.

Each policy network takes as input a three-channel RGB $200 \times 200$ image, downscaled from $640 \times 480$ in the dataset, applies three convolutional layers with ReLU activations followed by two fully-connected dense layers reducing to $64$ and then eight real-valued components. An action is represented by 3D translations and the opening angles of the left and right arm grippers.

We investigate how well the segmentation provided by DDO corresponds to expert annotations, when applied to demonstrations of the three surgical tasks. Figure 5 shows a representative sample of 10 trajectories from each task, with each time step colored by the most likely option to be active at that time. Human boundary annotations are marked in ×. We quantify the match between the manual and automatic annotation by the fraction of option boundaries that have exactly one human annotation in a 300 ms window around them. By this metric, DDO obtains 72% accuracy, while random guessing gets only 14%. These results suggest that DDO succeeds in learning some latent structure of the task.

## 6 Discussion

In this paper we presented the DDO algorithm that discovers parametrized options from a set of demonstration trajectories, and can be used recursively to discover multi-level hierarchies. Our results demonstrate that the discovered options accelerate learning in RL problems.

An important point for discussion is setting the hyper-parameters of DDO, namely, (1) the number of options at each level of the hierarchy, (2) how many RL episodes to wait before applying DDO, and (3) how to design the parametrization for the options. In our experiments, for (1), we tuned the number of options based on observing the RL agent's performance using those options. For (2), we ran the RL agent with only the actions, and observed the time step when its greedy policy started accumulating above-random rewards. We empirically found that this was a good stopping point to sample the greedy policy and apply DDO. Finally, we selected the option parametrization to be similar in architecture to the Q-Networks used on the domains. More principled methods for selecting all of these hyperparameters merit further research.

In some of our experiments we train the meta-control policy during option discovery. When we subsequently train a higher-level policy using the discovered options to augment the action space, we initialize the high-level $Q$ function to 0, rather than reusing any pre-training from the discovery stage. Our algorithm outperforms the baseline despite the handicap of this $Q$-function reset. A method for initializing the $Q$ function from discovery-stage pre-trained values, without introducing too much bias, could accelerate learning further.

Even with $Q$-function reset, options trained jointly with a meta-control policy accelerate the high-level learning stage more than options trained with a fixed uniform meta-control policy. However, when more levels are added to the hierarchy beyond the lowest two, our experiments suggest that the relative advantage is reversed: learning is faster when using lowest-level options that were trained with a uniform meta-control policy at the first stage. This preliminary finding indicates that option discovery with a simplified meta-control policy may not hurt performance, and perhaps even help.

Our derivation and algorithms apply to continuous as well as discrete action spaces, although in this paper we experimented mainly with the latter. A meta-control policy over a continuous action space augmented with a finite set of options can be represented by a hybrid network, that outputs parameters for the continuous action distribution, the discrete option distribution, and a bit to choose between them. The effectiveness of our approach in such domains remains an open question.

## Acknowledgements

This research was performed at the AUTOLAB at UC Berkeley in affiliation with the Berkeley AI Research (BAIR) Lab, the Real-Time Intelligent Secure Execution (RISE) Lab, and the CITRIS "People and Robots" (CPAR) Initiative. The authors were supported in part by the U.S. National Science Foundation under NRI Award IIS-1227536: Multilateral Manipulation by Human-Robot Collaborative Systems and the Berkeley Deep Drive (BDD) Program, by DHS Award HSHQDC-16-3-00083, NSF CISE Expeditions Award CCF-1139158, and donations and gifts from Siemens, Google, Cisco, Autodesk, IBM, Ant Financial, Amazon Web Services, CapitalOne, Ericsson, GE, Huawei, Intel, Microsoft and VMware.

Any opinions, findings, and conclusions or recommendations expressed in this material are those of the author(s) and do not necessarily reflect the views of the Sponsors.

## References

[1] Richard S. Sutton, Doina Precup, and Satinder P. Singh. Between MDPs and semi-MDPs: A framework for temporal abstraction in reinforcement learning. *AI*, 112(1-2):181–211, 1999.

[2] Tejas D Kulkarni, Karthik Narasimhan, Ardavan Saeedi, and Josh Tenenbaum. Hierarchical deep reinforcement learning: Integrating temporal abstraction and intrinsic motivation. In *NIPS*, pages 3675–3683, 2016.

[3] Nicolas Heess, Greg Wayne, Yuval Tassa, Timothy Lillicrap, Martin Riedmiller, and David Silver. Learning and transfer of modulated locomotor controllers. *arXiv preprint arXiv:1610.05182*, 2016.

[4] Pierre-Luc Bacon, Jean Harb, and Doina Precup. The option-critic architecture. *arXiv preprint arXiv:1609.05140*, 2016.

[5] Christian Daniel, Gerhard Neumann, and Jan Peters. Hierarchical relative entropy policy search. In *AISTATS*, pages 273–281, 2012.

[6] Aravind S Lakshminarayanan, Ramnandan Krishnamurthy, Peeyush Kumar, and Balaraman Ravindran. Option discovery in hierarchical reinforcement learning using spatio-temporal clustering. *arXiv preprint arXiv:1605.05359*, 2016.

[7] Mandana Hamidi, Prasad Tadepalli, Robby Goetschalckx, and Alan Fern. Active imitation learning of hierarchical policies. In *IJCAI*, pages 3554–3560, 2015.

[8] Hung Hai Bui, Svetha Venkatesh, and Geoff West. Policy recognition in the abstract hidden Markov model. *JAIR*, 17:451–499, 2002.

[9] Ronald Edward Parr. *Hierarchical control and learning for Markov decision processes*. PhD thesis, UNIVERSITY of CALIFORNIA at BERKELEY, 1998.

[10] Andrew G. Barto and Sridhar Mahadevan. Recent advances in hierarchical reinforcement learning. *Discrete Event Dynamic Systems*, 13(1-2):41–77, 2003.

[11] George Konidaris, Scott Kuindersma, Roderic A. Grupen, and Andrew G. Barto. Robot learning from demonstration by constructing skill trees. *IJRR*, 31(3):360–375, 2012.

[12] Sanjay Krishnan, Animesh Garg, Richard Liaw, Brijen Thananjeyan, Lauren Miller, Florian T Pokorny, and Ken Goldberg. SWIRL: A sequential windowed inverse reinforcement learning algorithm for robot tasks with delayed rewards. In *WAFR*, 2016.

[13] Pierre Sermanet, Kelvin Xu, and Sergey Levine. Unsupervised perceptual rewards for imitation learning. *arXiv preprint arXiv:1612.06699*, 2016.

[14] Matthew M Botvinick. Hierarchical models of behavior and prefrontal function. *Trends in cognitive sciences*, 12(5):201–208, 2008.

[15] Matthew M Botvinick, Yael Niv, and Andrew C Barto. Hierarchically organized behavior and its neural foundations: A reinforcement learning perspective. *Cognition*, 113(3):262–280, 2009.

[16] Alec Solway, Carlos Diuk, Natalia Córdova, Debbie Yee, Andrew G Barto, Yael Niv, and Matthew M Botvinick. Optimal behavioral hierarchy. *PLOS Comput Biol*, 10(8):e1003779, 2014.

[17] Jeffrey M Zacks, Christopher A Kurby, Michelle L Eisenberg, and Nayiri Haroutunian. Prediction error associated with the perceptual segmentation of naturalistic events. *Journal of Cognitive Neuroscience*, 23(12):4057–4066, 2011.

[18] Andrew Whiten, Emma Flynn, Katy Brown, and Tanya Lee. Imitation of hierarchical action structure by young children. *Developmental science*, 9(6):574–582, 2006.

[19] Rodney Brooks. A robust layered control system for a mobile robot. *IEEE journal on robotics and automation*, 2(1):14–23, 1986.

[20] Peter Dayan and Geoffrey E. Hinton. Feudal reinforcement learning. In *NIPS*, pages 271–278, 1992.

[21] Bernhard Hengst. Discovering hierarchy in reinforcement learning with HEXQ. In *ICML*, volume 2, pages 243–250, 2002.

[22] J Zico Kolter, Pieter Abbeel, and Andrew Y Ng. Hierarchical apprenticeship learning with application to quadruped locomotion. In *NIPS*, volume 20, 2007.

[23] George Konidaris and Andrew G Barto. Building portable options: Skill transfer in reinforcement learning. In *IJCAI*, volume 7, pages 895–900, 2007.

[24] Leslie Pack Kaelbling. Hierarchical learning in stochastic domains: Preliminary results. In *ICML*, pages 167–173, 1993.

[25] Manfred Huber and Roderic A. Grupen. A feedback control structure for on-line learning tasks. *Robotics and Autonomous Systems*, 22(3-4):303–315, 1997.

[26] Pierre-Luc Bacon and Doina Precup. Learning with options: Just deliberate and relax. In *NIPS Bounded Optimality and Rational Metareasoning Workshop*, 2015.

[27] Richard Liaw, Sanjay Krishnan, Animesh Garg, Daniel Crankshaw, Joseph E Gonzalez, and Ken Goldberg. Composing meta-policies for autonomous driving using hierarchical deep reinforcement learning. 2017.

[28] Ronald Parr and Stuart J. Russell. Reinforcement learning with hierarchies of machines. In *NIPS*, pages 1043–1049, 1997.

[29] Thomas G Dietterich. Hierarchical reinforcement learning with the MAXQ value function decomposition. *JAIR*, 13:227–303, 2000.

[30] Sebastian Thrun and Anton Schwartz. Finding structure in reinforcement learning. In *NIPS*, pages 385–392, 1994.

[31] Amy McGovern and Andrew G. Barto. Automatic discovery of subgoals in reinforcement learning using diverse density. In *ICML*, pages 361–368, 2001.

[32] Ishai Menache, Shie Mannor, and Nahum Shimkin. Q-cut—dynamic discovery of sub-goals in reinforcement learning. In *ECML*, pages 295–306. Springer, 2002.

[33] Özgür Şimşek and Andrew G Barto. Using relative novelty to identify useful temporal abstractions in reinforcement learning. In *ICML*, page 95. ACM, 2004.

[34] Martin Stolle. *Automated discovery of options in reinforcement learning*. PhD thesis, McGill University, 2004.

[35] Sanjay Krishnan, Animesh Garg, Sachin Patil, Colin Lea, Gregory Hager, Pieter Abbeel, and Ken Goldberg. Transition state clustering: Unsupervised surgical trajectory segmentation for robot learning. In *ISRR*, 2015.

[36] George Konidaris and Andrew G Barto. Skill discovery in continuous reinforcement learning domains using skill chaining. In *NIPS*, pages 1015–1023, 2009.

[37] Kfir Y Levy and Nahum Shimkin. Unified inter and intra options learning using policy gradient methods. In *European Workshop on Reinforcement Learning*, pages 153–164. Springer, 2011.

[38] Tim Genewein, Felix Leibfried, Jordi Grau-Moya, and Daniel Alexander Braun. Bounded rationality, abstraction, and hierarchical decision-making: An information-theoretic optimality principle. *Frontiers in Robotics and AI*, 2:27, 2015.

[39] Roy Fox, Michal Moshkovitz, and Naftali Tishby. Principled option learning in Markov decision processes. *arXiv preprint arXiv:1609.05524*, 2016.

[40] Anders Jonsson and Vicenç Gómez. Hierarchical linearly-solvable Markov decision problems. *arXiv preprint arXiv:1603.03267*, 2016.

[41] Carlos Florensa, Yan Duan, and Pieter Abbeel. Cstochastic neural networks for hierarchical reinforcement learning. In *ICLR*, 2017.

[42] Sahil Sharma, Aravind S. Lakshminarayanan, and Balaraman Ravindran. Learning to repeat: Fine grained action repetition for deep reinforcement learning. In *ICLR*, 2017.

[43] Christian Daniel, Herke Van Hoof, Jan Peters, and Gerhard Neumann. Probabilistic inference for determining options in reinforcement learning. *Machine Learning*, 104(2-3):337–357, 2016.

[44] Tom Schaul, Daniel Horgan, Karol Gregor, and David Silver. Universal value function approximators. In *ICML*, pages 1312–1320, 2015.

[45] Christopher JCH Watkins and Peter Dayan. Q-learning. *Machine learning*, 8(3-4):279–292, 1992.

[46] Richard S Sutton and Andrew G Barto. *Reinforcement learning: An introduction*, volume 1. MIT press Cambridge, 1998.

[47] Volodymyr Mnih, Koray Kavukcuoglu, David Silver, Andrei A Rusu, Joel Veness, Marc G Bellemare, Alex Graves, Martin Riedmiller, Andreas K Fidjeland, Georg Ostrovski, et al. Human-level control through deep reinforcement learning. *Nature*, 518(7540):529–533, 2015.

[48] Ruslan Salakhutdinov, Sam Roweis, and Zoubin Ghahramani. Optimization with EM and expectation-conjugate-gradient. In *ICML*, pages 672–679, 2003.

[49] Geoffrey McLachlan and Thriyambakam Krishnan. *The EM algorithm and extensions*, volume 382. John Wiley & Sons, 2007.

[50] Leonard E Baum. An equality and associated maximization technique in statistical estimation for probabilistic functions of markov processes. *Inequalities*, 3:1–8, 1972.

[51] Jakub Sygnowski and Henryk Michalewski. Learning from the memory of Atari 2600. *arXiv preprint arXiv:1605.01335*, 2016.

[52] Yixin et al. Gao. The JHU-ISI gesture and skill assessment dataset (jigsaws): A surgical activity working set for human motion modeling. In *Medical Image Computing and Computer-Assisted Intervention (MICCAI)*, 2014.

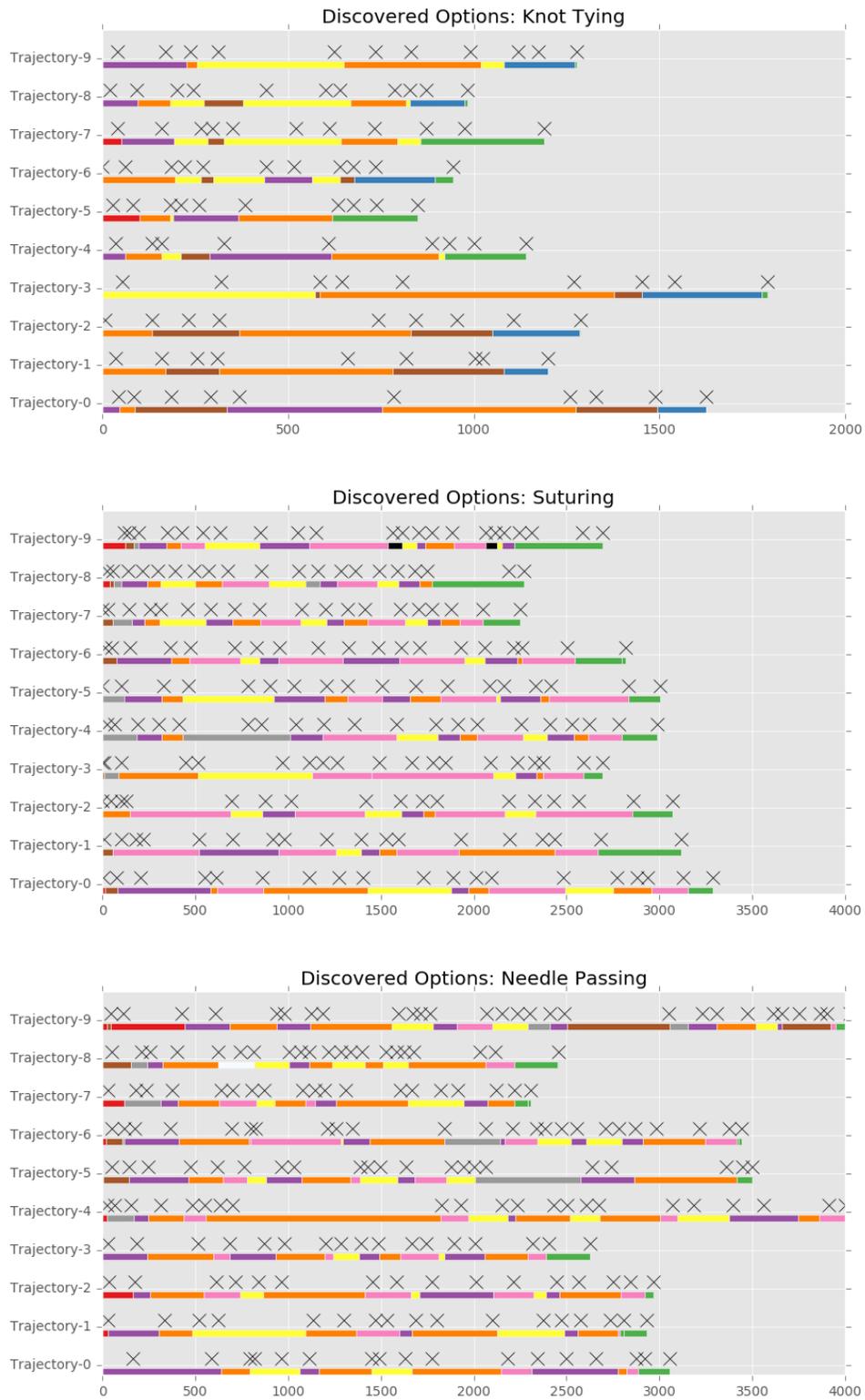

Figure 5: Segmentation of human demonstrations in three surgical tasks. Each line represents a trajectory, with segment color indicating the option inferred by DDO as most likely to be active at that time. Human annotations are marked as × above the segments. Automated segmentation achieves a good alignment with human annotation — 72% of option boundaries have exactly one annotation in a 300 ms window around them.

# A Forward-Backward Algorithm

Despite the exponential domain size of the latent variable $\zeta$, Expectation-Gradient for trajectories allows us to decompose the posterior $\mathbb{P}_\theta(\zeta|\xi)$ and only concern ourselves with each marginal posterior separately. These marginal posteriors can be computed by a forward-backward dynamic programming algorithm, similar to Baum-Welch [50].

We compute the likelihood of a trajectory prefix

$$\phi_t(h) = \mathbb{P}_\theta(s_0, a_0, \ldots, s_t, h_t{=}h),$$

using the forward recursion

$$\phi_0(h) = p_0(s_0)\eta(h|s_0)$$

and for $0 \leqslant t < T - 1$

$$\phi_{t+1}(h') = \sum_{h \in \mathcal{H}} \phi_t(h)\pi_h(a_t|s_t)p(s_{t+1}|s_t, a_t)(\psi_h(s_{t+1})\eta(h'|s_{t+1}) + (1 - \psi_h(s_{t+1}))\delta_{h=h'})$$

$$= \sum_{h \in \mathcal{H}} \left(\phi_t(h)\pi_h(a_t|s_t)p(s_{t+1}|s_t, a_t)\psi_h(s_{t+1})\right)\eta(h'|s_{t+1})$$

$$+ \phi_t(h')\pi_{h'}(a_t|s_t)p(s_{t+1}|s_t, a_t)(1 - \psi_{h'}(s_{t+1})).$$

We similarly compute the likelihood of a trajectory suffix

$$\omega_t(h) = \mathbb{P}_\theta(a_t, s_{t+1}, \ldots, s_T|s_t, h_t{=}h),$$

using the backward recursion

$$\omega_{T-1}(h) = \pi_h(a_{T-1}|s_{T-1})p(s_T|s_{T-1}, a_{T-1})$$

and for $0 \leqslant t < T - 1$

$$\omega_t(h) = \pi_h(a_t|s_t)p(s_{t+1}|s_t, a_t)\left(\psi_h(s_{t+1})\sum_{h' \in \mathcal{H}} \eta(h'|s_{t+1})\omega_{t+1}(h') + (1 - \psi_h(s_{t+1}))\omega_{t+1}(h)\right).$$

When we know that the trajectory ends in option termination, we also need to multiply $\omega_{T-1}(h)$ by $\psi_h(s_T)$, and add to the log-likelihood gradient the term $\nabla_\theta \log \psi_h(s_T)$ with weight $u_{T-1}$.

We can compute our target likelihood using any $0 \leqslant t < T$

$$\mathbb{P}_\theta(\xi) = \sum_{h \in \mathcal{H}} \mathbb{P}_\theta(\xi, h_t{=}h) = \sum_{h \in \mathcal{H}} \phi_t(h)\omega_t(h).$$

Recall that we define

$$u_t(h) = \mathbb{P}_\theta(h_t{=}h|\xi)$$
$$v_t(h) = \mathbb{P}_\theta(b_t{=}1, h_t{=}h|\xi)$$
$$w_t(h) = \mathbb{P}_\theta(h_t{=}h, b_{t+1}{=}0|\xi).$$

The marginal posteriors are

$$u_t(h) = \frac{1}{\mathbb{P}(\xi)}\phi_t(h)\omega_t(h)$$

$$v_0(h) = u_0(h)$$

and for $0 \leqslant t < T - 1$

$$v_{t+1}(h') = \frac{1}{\mathbb{P}(\xi)}\left(\sum_{h \in \mathcal{H}} \phi_t(h)\pi_h(a_t|s_t)p(s_{t+1}|s_t, a_t)\psi_h(s_{t+1})\right)\eta(h'|s_{t+1})\omega_{t+1}(h')$$

$$w_t(h) = \frac{1}{\mathbb{P}(\xi)}\phi_t(h)\pi_h(a_t|s_t)p(s_{t+1}|s_t, a_t)(1 - \psi_h(s_{t+1}))\omega_{t+1}(h).$$

Note that the constant $p_0(s_0)\prod_{t=0}^{T-1} p(s_{t+1}|s_t, a_t)$ is cancelled out in these normalizations. This allows us to omit these terms during the forward-backward algorithm, which can thus be applied without any knowledge of the dynamics.

# B Supplemental Experiments

This section includes a number of simplified examples to convey intuition of why options can improve RL performance.

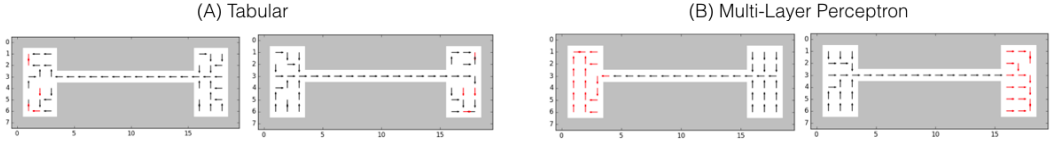

Figure 6: Two primitives learned with (a) a tabular representation; (B) multi-layer perceptron. Termination probability are visualized in red.

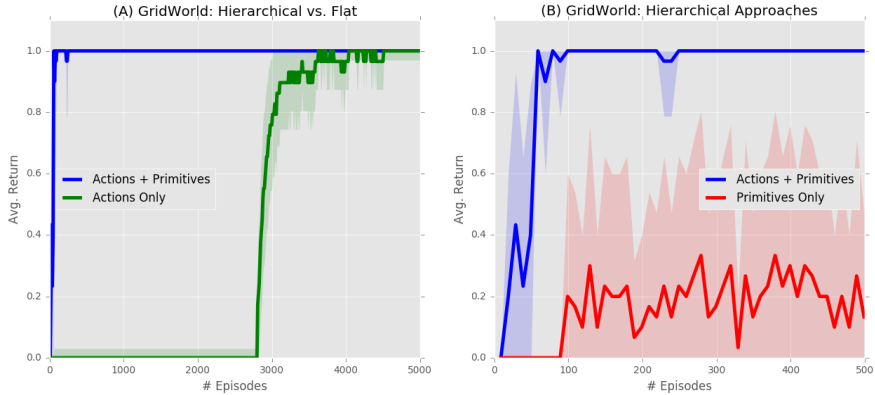

Figure 7: Given a new start and end state, we plot the value after $t$ episodes of RL with a DQN. The learned options can facilitate more efficient RL as the agent can leverage previously learned skills.

### GridWorld: Supervised

In the first scenario, we construct a 10x20 GridWorld environment with two rooms with a 30% probability of random action. We generate 10 demonstrations from randomly chosen start and end points situated in different rooms using Value Iteration. We apply the inference algorithm to learn 2 options (Figure 6). Not surprisingly, there is one option policy that exits the left room and moves to the right, and another that exits the right room and moves to the left. The termination probability is visualized in red. Figure 6 illustrates the flexibility of the proposed inference algorithm, namely, we can apply it to a tabular (fully parameterized) representation as well as a neural network representation (multi-layer perceptron MLP). We can see that the MLP generalizes to unseen states while still being expressive enough to model the geometry of the two rooms. This is especially true for the parameterization of the termination probabilities. The MLP learns to terminate the primitive in each of the rooms while the tabular representation does not make that generalization.

Given a new start and end goal, we can use reinforcement learning to find a policy. We add a reward function to the two-rooms environment which gives a reward of +1 to the agent if it reaches the goal in 20 time-steps. We evaluate the extent to which the options learned reduce exploration to find this policy. We use the MLP options described in the previous experiments. We use a Deep Q Network (DQN) as the RL algorithm. We compare three approaches: (1) DQN using the primitive actions, (2) DQN using only the options, and (3) DQN using an augmented action space of the primitive actions and options. Figure 7 plots the estimated value of each policy from 30 trials. The DQN augmented with action and options converges to the maximum reward in 100 episodes, whereas the DQN over just the primitive actions requires 4600. Figure 7b illustrates why this is likely the case. When we compare the options-only to the approach with both options and actions, we see that the options-only approach is able to quickly receive a positive reward but not always able to reach the goal. This early positive reward signal can greatly speed up the convergence.

### GridWorld: Demonstrations and Initialization

Next, we discuss some additional details about the DDO algorithm. First, we explore the number and quality of demonstrations needed to learn viable options. As before, we consider an MLP representation for the options. Figure 8a varies the environment noise in the GridWorld and plots the number of episodes to convergence (defined as a reward of 1.0 on 10 consecutive trials). For the value iteration demonstrator, a fairly small number of demonstrations is required to learn useful options. This number is relatively robust to environments with small amounts of stochasticity. On the other hand, Figure 8b plots "supervisor" noise in the demonstrations. We randomly corrupt a fraction of the taken actions from the supervisor. For this type of noise, a significantly larger number of demonstrations is required to see the same gains in performance as the infallable supervisor.

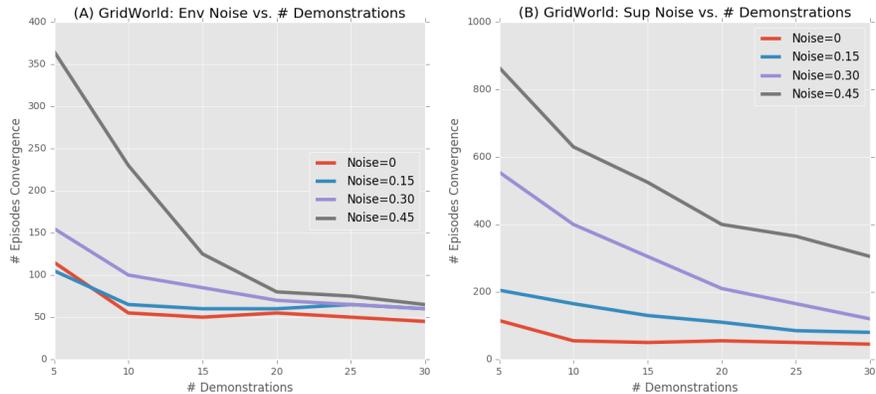

Figure 8: We plot the number of demonstrations needed to achieve a certain convergence rate as a function of environment and supervisor noise.

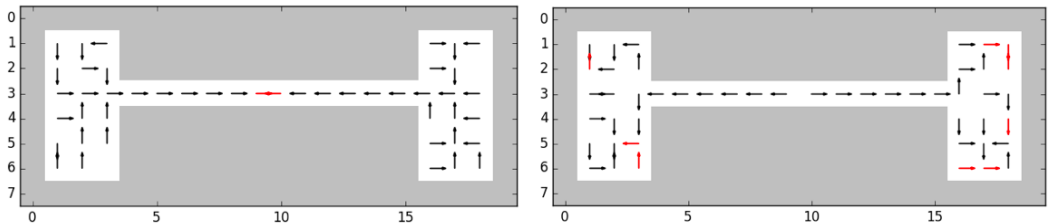

Figure 9: Initializing by pre-clustering can result in qualitatively different results for the same environment.

Another further point of discussion is the initialization of the DDO algorithm. As in clustering algorithms, one must initialize the model in a way that breaks any symmetry. The initialization approach can affect the solutions that arise. In particular, we can use an approach to initialize where the state action tuples are first clustered and then initial models are trained on each cluster. This is similar to techniques used to initialize k-means. When we use this initialization for the tabular representation, this leads to a qualitatively different set of primitives for the two rooms environment than in the previous result (Figure 9). There is one option that leads the agent out of each room to the center and another that takes the agent from the center to each room. One could argue that this set of primitives is easier to plan with as the termination condition is far simpler than in the previous result.

**GridWorld: Exploration**

Next, we show how within a task options can be used to improve the convergence of an RL algorithm. We construct a 10x25 GridWorld environment with three rooms with a 30% probability of random action. The agent starts in the leftmost room, has to reach a point in the second room, and then progress to the third room. The agent receives a reward of 0.5 when it reaches the first goal, and receives a reward of 0.5 when it reaches the second goal only if it had reached the first goal. Figure 10 illustrates the domain and the results.

We apply RL with a DQN for 5000 episodes. This roughly corresponds to the agent reliably reaching the first goal. Then, we apply the DDO algorithm to learn two options. For future iterations, we augment the DQN with the learned options. Figure 10b illustrates how this can improve the convergence. The agent can use the options as a fixed subroutine that can take it to the first goal, and then from that point, it can try further actions. This focuses the exploration towards searching for the goal in the second room.

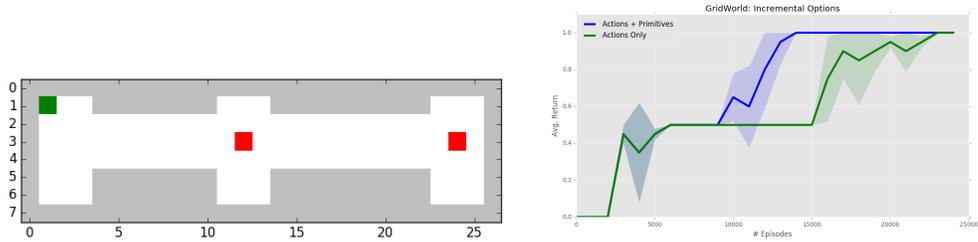

Figure 10: In a three-room gridworld, we use DDO to incrementally construct options during Reinforcement Learning after 5000 initial episodes. The future iterations can use the options and this significantly speeds up convergence.

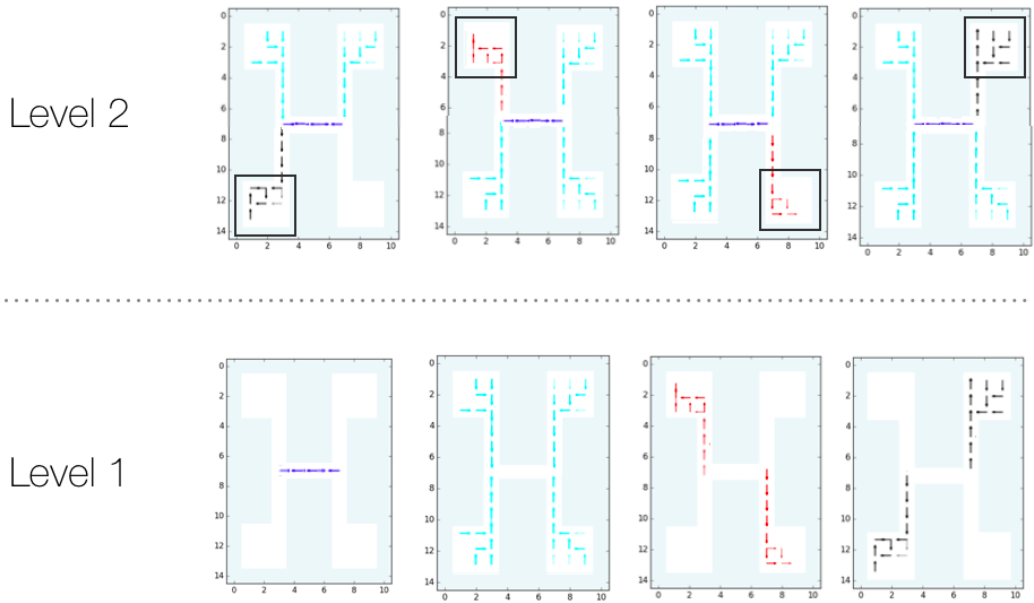

Figure 11: In a four-room gridworld, discovering options with a trainable meta-control policy may lead to options that generalize worse, due to the ability of the higher level to compose them in sophisticated ways. Here three options rather than two are needed to reach a specific room.

**State-Dependent: Four Rooms**

To understand why the richer high-level meta policy potentially hurts in the presented Four-Room GridWorld example, we visualize the learned options in Figure 11. One of the learned options is an overly specific routine traversing the small corridor. This leads to a more complex high-level policy that composes three options to go to each room instead of two. However, we caution that this is not definitive evidence that simplifying the high-level policy improves option discovery. It suggests that a richer high-level policy potentially requires more regularization and more accurate initialization.

**Atari Neural Network Architecture**

The neural network used in the Atari domain to represent options (control and termination policies) and meta-control policies, is depicted in Figure 12.

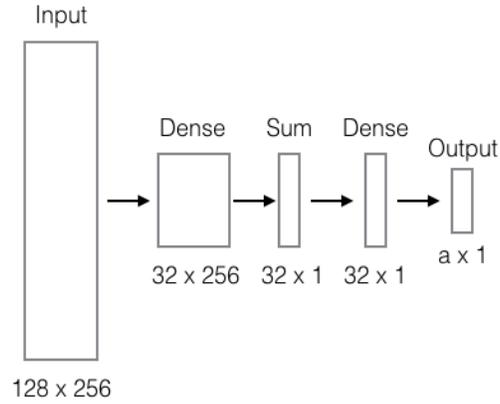

Figure 12: Network architecture for the policies and DQN in the Atari experiments. The output distribution is the softmax of the network output vector of length $|\mathcal{A}|$ for control policies and 2 for termination.

## C  Additional Considerations

Attempting to fit the hierarchical generative model to demonstrations makes an implicit assumption that there is useful structure to discover in the first place. Hierarchy assumes that the meta-control policy can identify, for each option $h$, a set of states where $h$ is advantageous, by setting $\eta(h|s)$ high; that the advantage of using the policy $\pi_h$, when used in such a state, is likely to persist to the next state; and that the termination policy can identify states where $h$ stops being advantageous, by setting $\psi_h(s)$ high. Under these assumptions, the meta-control policy benefits from only attending to slowly changing state features, while each option benefits from only attending to local state features.

A key consideration is the expressive power of the parametrization of $\pi_h$, $\psi_h$ and $\eta$, and its effect on the "life expectancy" of options, i.e. the expected time until one terminates. Any policy $\pi(a|s)$ can be represented as a single low-level option that never terminates, if it is expressive enough. At the other extreme, by including single-action options for each $a \in \mathcal{A}$, having $\pi_{h_a}(a'|s) = \delta_{a,a'}$ and immediate termination $\psi_{h_a}(s) = 1$, the meta-control policy can implement any policy directly with $\eta(h_a|s) = \pi(a|s)$.